\documentclass[11pt,a4paper]{article}
\usepackage[hyperref]{acl2019}
\usepackage{times}
\usepackage{latexsym}
\usepackage{graphicx}
\usepackage{enumitem}
\usepackage{amsmath}
\usepackage{amssymb}
\usepackage{bm}
\usepackage{algorithm}
\usepackage{algpseudocode}
\usepackage{subcaption}
\usepackage{hhline}

\usepackage{url}

\aclfinalcopy 

\title{Neural Aspect and Opinion Term Extraction with Mined Rules as \\ Weak Supervision}

\author{Hongliang Dai \\
	Department of CSE, HKUST \\
	{\tt hdai@cse.ust.hk} \\\And
	Yangqiu Song \\
	Department of CSE, HKUST \\
	{\tt yqsong@cse.ust.hk} \\}

\date{}

\begin{document}
\maketitle
\begin{abstract}
Lack of labeled training data is a major bottleneck for neural network based aspect and opinion term extraction on product reviews. 
To alleviate this problem, we first propose an algorithm to automatically mine extraction rules from existing training examples based on dependency parsing results. The mined rules are then applied to label a large amount of auxiliary data. Finally, we study training procedures to train a neural model which can learn from both the data automatically labeled by the rules and a small amount of data accurately annotated by human.
Experimental results show that although the mined rules themselves do not perform well due to their limited flexibility, the combination of human annotated data and rule labeled auxiliary data can improve the neural model and allow it to achieve performance better than or comparable with the current state-of-the-art.
\end{abstract}

\section{Introduction}

There are two types of words or phrases in product reviews (or reviews for services, restaurants, etc., we use ``product reviews'' throughout the paper for convenience) that are of particular importance for opinion mining: those that describe a product's properties or attributes; and those that correspond to the reviewer's sentiments towards the product or an aspect of the product \cite{hu2004mining,liu2012sentiment,qiu2011opinion,vivekanandan2014aspect}. The former are called aspect terms, and the latter are called opinion terms. For example, in the sentence ``The speed of this laptop is incredible,'' ``speed'' is an aspect term, and ``incredible'' is an opinion term. The task of aspect and opinion term extraction is to extract the above two types of terms from product reviews.
	
Rule based approaches \cite{qiu2011opinion,liu2016improving} and learning based approaches \cite{jakob2010extracting,wang2016recursive} are two major approaches to this task. Rule based approaches usually use manually designed rules based on the result of dependency parsing to extract the terms. An advantage of these approaches is that the aspect or opinion terms whose usage in a sentence follows some certain patterns can always be extracted. However, it is labor-intensive to design rules manually. It is also hard for them to achieve high performance due to the variability and ambiguity of natural language.

Learning based approaches model aspect and opinion term extraction as a sequence labeling problem. While they are able to obtain better performance, they also suffer from the problem that significant amounts of labeled data must be used to train such models to reach their full potential, especially when the input features are not manually designed. Otherwise, they may even fail in very simple test cases (see Section \ref{sec:casestudy} for examples).
	
	In this paper, to address above problems, we first use a rule based approach to extract aspect and opinion terms from an auxiliary set of product reviews, which can be considered as inaccurate annotation. These rules are automatically mined from the labeled data based on dependency parsing results. Then, we propose a BiLSTM-CRF (Bi-directional LSTM-Conditional Random Field) based neural model for aspect and opinion term extraction. 
	This neural model is trained with both the human annotated data as ground truth supervision and the rule annotated data as weak supervision. 
	We name our approach RINANTE (\textbf{R}ule \textbf{I}ncorporated \textbf{N}eural \textbf{A}spect and Opi\textbf{n}ion \textbf{T}erm \textbf{E}xtraction). 

	We conduct experiments on three SemEval datasets that are frequently used in existing aspect and opinion term extraction studies. The results show that the performance of the neural model can be significantly improved by training with both the human annotated data and the rule annotated data. 
	
	Our contributions are summarized as follows.
	
	\begin{itemize}
		\item We propose to improve the effectiveness of a neural aspect and opinion term extraction model by training it with not only the human labeled data but also the data automatically labeled by rules.
		\item We propose an algorithm to automatically mine rules based on dependency parsing and POS tagging results for aspect and opinion term extraction.
		\item We conduct comprehensive experiments to verify the effectiveness of the proposed approach.
	\end{itemize}
	
	Our code is available at \url{https://github.com/HKUST-KnowComp/RINANTE}.

\section{Related Work}
    There are mainly three types of approaches for aspect and opinion term extraction: rule based approaches, topic modeling based approaches, and learning based approaches.
    
    A commonly used rule based approach is to extract aspect and opinion terms based on dependency parsing results \cite{zhuang2006movie,qiu2011opinion}. A rule in these approaches usually involves only up to three words in a sentence \cite{qiu2011opinion}, which limits its flexibility. It is also labor-intensive to design the rules manually. \citet{liu2015automated} propose an algorithm to select some rules from a set of previously designed rules, so that the selected subset of rules can perform extraction more accurately. However, different from the rule mining algorithm used in our approach, it is unable to discover rules automatically.

	Topic modeling approaches \cite{lin2009joint,brody2010unsupervised,mukherjee2012aspect} are able to get coarse-grained aspects such as \textit{food}, \textit{ambiance}, \textit{service} for restaurants, and provide related words. However, they cannot extract the exact aspect terms from review sentences.
	
Learning based approaches extract aspect and opinion terms by labeling each word in a sentence with BIO (Begin, Inside, Outside) tagging scheme \cite{ratinov2009design}. Typically, they first obtain features for each word in a sentence, then use them as the input of a CRF to get better sequence labeling results \cite{jakob2010extracting,wang2016recursive}. Word embeddings are commonly used features, hand-crafted features such as POS tag classes and chunk information can also be combined to yield better performance \cite{liu2015fine,yin2016unsupervised}. For example, \citet{wang2016recursive} construct a recursive neural network based on the dependency parsing tree of a sentence with word embeddings as input. The output of the neural network is then fed into a CRF. \citet{xu2018double} use a CNN model to extract aspect terms. They find that using both general-purpose and domain-specific word embeddings improves the performance.
	
Our approach exploits unlabeled extra data to improve the performance of the model. This is related to semi-supervised learning and transfer learning. Some methods allow unlabeled data to be used in sequence labeling. For example, \citet{jiao2006semi} propose semi-supervised CRF, \citet{zhang2017semi} propose neural CRF autoencoder. Unlike our approach, these methods do not incorporate knowledge about the task while using the unlabeled data.
\citet{yang2017transfer} propose three different transfer learning architectures that allow neural sequence tagging models to learn from both the target task and a different but related task. Different from them, we improve performance by utilizing the output of a rule based approach for the same problem, instead of another related task. 

Our approach is also related to the use of weakly labeled data \cite{craven1999constructing}, and is similar to the distant supervision approach used in relation extraction \cite{mintz2009distant}.

\section{RINANTE}
\label{sec:rinante}
	
	In this section, we introduce our approach RINANTE in detail.
	Suppose we have a human annotated dataset $D_l$ and an auxiliary dataset $D_a$. $D_l$ contains a set of product reviews, each with all the aspect and opinion terms in it labeled. $D_a$ only contains a set of unlabeled product reviews. The reviews in $D_l$ and $D_a$ are all for a same type or several similar types of products. Usually, the size of $D_a$ is much larger than $D_l$. Then, RINANTE consists of the following steps.

	\begin{itemize}
		\item[1.] Use $D_l$ to mine a set of aspect extraction rules $R_a$ and a set of opinion extraction rules $R_o$ with a rule mining algorithm.
		\item[2.] Use the mined rules $R_a$ and $R_o$ to extract terms for all the reviews in $D_a$, which can then be considered a weakly labeled dataset $D_a'$.
		\item[3.] Train a neural model with $D_l$ and $D_a'$. The trained model can be used on unseen data.
	\end{itemize}
	
	Next, we introduce the rule mining algorithm used in Step 1 and the neural model in Step 3.
	
	\subsection{Rule Mining Algorithm}
	
	\begin{figure}
		\centering
		\includegraphics[width=55mm]{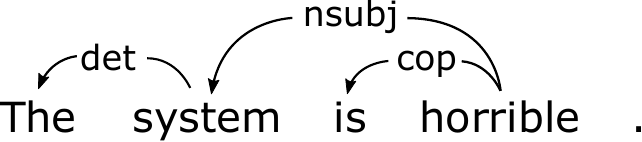}
		\caption{The dependency relations between the words in sentence ``The system is horrible.'' Each edge is a relation from the governor to the dependent.}
		\label{fig:dep-parse}
	\end{figure}
	
	We mine aspect and opinion term extraction rules that are mainly based on the dependency relations between words, since their effectiveness has been validated by existing rule based approaches \cite{zhuang2006movie,qiu2011opinion}. 
	
	We use $(rel, w_g, w_d)$ to denote that the dependency relation $rel$ exists between the word $w_g$ and the word $w_d$, where $w_g$ is the governor and $w_d$ is the dependent.
	An example of the dependency relations between different words in a sentence is given in Figure \ref{fig:dep-parse}. In this example, ``system'' is an aspect term, and ``horrible'' is an opinion term.
	A commonly used rule to extract aspect terms is $(nsubj, O, noun^*)$, where we use $O$ to represent a pattern that matches any word that belongs to a predefined opinion word vocabulary; $noun^*$ matches any noun word and the $^*$ means that the matched word is output as the aspect word. With this rule, the aspect term ``system'' in the example sentence can be extracted if the opinion term ``horrible'' can be matched by $O$.
	
	The above rule involves two words. In our rule mining algorithm, we only mine rules that involve no more than three words, because rules that involve many words may contribute very little to recall but are computationally expensive to mine. Moreover, determining their effectiveness requires a lot more labeled data since such patterns do not occur frequently.
	Since the aspect term extraction rule mining algorithm and the opinion term extraction rule mining algorithm are similar, we only introduce the former in detail. The algorithm contains two main parts: 1) Generating rule candidates based on a training set; 2) Filtering the rule candidates based on their effectiveness on a validation set. 
	
	The pseudocode for generating aspect term extraction rule candidates is in Algorithm~\ref{alg:gencand}. In Algorithm~\ref{alg:gencand}, $s_i.aspect\_terms$ is a list of the manually annotated aspect terms in sentence $s_i$, $s_i.deps$ is the list of the dependency relations obtained after performing dependency parsing. $list1$ and $list2$ contain the possible term extraction patterns obtained from each sentence that involve two and three words, respectively.

	\begin{algorithm}[t]
		\caption{Aspect term extraction rule candidate generation}
		\label{alg:gencand}
		
		\begin{itemize}[noitemsep,topsep=0pt,leftmargin=11.5mm]
			\item[\textbf{Input:}] A set of sentences $S_t$ with all aspect terms extracted; integer $T$.
		\end{itemize}
		
		\textbf{Output:} $RC$
		\begin{algorithmic}[1]
			\State Initialize $list1$, $list2$ as empty lists
			\For{$s_i \in S_t$}
			\For{$a_i \in s_i.aspect\_terms$}
			\State $D1 = \text{RelatedS1Deps}(a_i, s_i.deps)$
			\State $D2 = \text{RelatedS2Deps}(a_i, s_i.deps)$
			\State $list1 \mathrel{+}= \text{PatternsFromS1Deps}(D1)$
			\State $list2 \mathrel{+}= \text{PatternsFromS2Deps}(D2)$
			\EndFor
			\EndFor
			\State $RC1=\text{FrequentPatterns}(list1, T)$
			\State $RC2=\text{FrequentPatterns}(list2, T)$
			\State $RC=RC1+RC2$
		\end{algorithmic}
	\end{algorithm}
	
	The function RelatedS1Deps on Line 4 returns a list of dependency relations. Either the governor or the dependent of each dependency relation in this list has to be a word in the aspect term. The function PatternsFromS1Deps is then used to get aspect term extraction patterns that can be obtained from the dependency relations in this list. Let $\text{POS}(w_d)$ be the POS tag of $w_d$; $ps(w)$ be a function that returns the word type of $w$ based on its POS tag, e.g., $noun$, $verb$, etc. Then for each $(rel, w_g, w_d)$, if $w_d$ is a word in the aspect term, PatternsFromS1Deps may generate the following patterns: $(rel,\allowbreak w_g, ps(w_d)^*)$, $(rel,\allowbreak \text{POS}(w_g),\allowbreak ps(w_d)^*)$ and $(rel,\allowbreak O,\allowbreak ps(w_d)^*)$. For example, for $(nsubj, \text{``horrible''},\allowbreak \text{``system''})$, it generates three patterns: $(nsubj, \text{``horrible''}, noun^*)$, $(rel, JJ, noun^*)$ and $(rel, O, noun^*)$. Note that $(rel, O, ps(w_d)^*)$ is only generated when $w_g$ belongs to a predefined opinion word vocabulary. Also, we only consider two types of words while extracting aspect terms: nouns and verbs, i.e., we only generate the above patterns when $ps(w_g)$ returns $noun$ or $verb$. The patterns generated when $w_g$ is the word in the aspect term are similar.
	
	The function RelatedS2Deps on Line 5 returns a list that contains pairs of dependency relations. The two dependency relations in each pair must have one word in common, and one of them is obtained with RelatedS1Deps. Afterwards, PatternsFromS2Deps generates patterns based on the dependency relation pairs. For example, the pair $\{(nsubj, \text{``like''}, \text{``I''}), (dobj, \text{``like''}, \text{``screen''})\}$ can be in the list returned by RelatedS2Deps, because ``like'' is the shared word, and $(dobj,\allowbreak \text{``like''},\allowbreak \text{``screen''})$ can be obtained with RelatedS1Deps since ``screen'' is an aspect term. A pattern generated based on this relation pair can be, e.g., $\{(nsubj, \text{``like''}, \text{``I''}), (dobj, \text{``like''}, noun^*)\}$. The operations of PatternsFromS2Deps is similar with PatternsFromS1Deps except patterns are generated based on two dependency relations.
	
	Finally, the algorithm obtains the rule candidates with the function FrequentPatterns, which counts the occurrences of the patterns and only return those that occur more than $T$ times. $T$ is a predefined parameter that can be determined based on the total number of sentences in $S$. $RC1$ and $RC2$ thus contains candidate patterns based on single dependency relations and dependency relation pairs, respectively. They are merged to get the final rule candidates list $RC$.
	
	\begin{algorithm}
		\caption{Aspect term extraction with mined rules}
		\label{alg:term-extract}
		
		\begin{itemize}[noitemsep,topsep=0pt,leftmargin=11.5mm]
			\item[\textbf{Input:}] Sentence $s$; rule pattern $r$; a set of phrases unlikely to be aspect terms $V_{fil}$.
		\end{itemize}
		
		\textbf{Output}: $A$
		\begin{algorithmic}[1]
			\State Initialize $A$ as en empty list.
			\For{$(rel, w_g, w_d) \in s.deps$}
			\If{$(rel, w_g, w_d)$ does not matches $r$}
			\State \textbf{continue}
			\EndIf
			\If{the governor of $r$ is the aspect word}
			\State $term=\text{TermFrom}(w_g)$
			\Else
			\State $term=\text{TermFrom}(w_d)$
			\EndIf
			\If{$term\notin V_{fil}$}
			\State $A.\text{add}(term)$
			\EndIf
			\EndFor
		\end{algorithmic}
	\end{algorithm}
	
	We still do not know the precision of the rule candidates obtained with Algorithm~\ref{alg:gencand}. Thus in the second part of our rule mining algorithm, for each rule candidate, we use it to extract aspect terms from another annotated set of review sentences (a validation set) and use the result to estimate its precision. Then we filter those whose precisions are less than a threshold $p$. The rest of the rules are the final mined rules. The algorithm for extracting aspect terms from a sentence $s$ with a rule pattern $r$ that contains one dependency relation is shown in Algorithm~\ref{alg:term-extract}. Since a rule pattern can only match one word in the aspect term, the function TermFrom in Algorithm~\ref{alg:term-extract} tries to obtain the whole term based on this matched seed word. Specifically, it simply returns the word $w_s$ when it is a verb. But when $w_s$ is a noun, it returns a noun phrase formed by the consecutive sequence of noun words that includes $w_s$. $V_{fil}$ is a set of phrases that are unlikely to be aspect terms. It includes the terms extracted with the candidate rules from the training set that are always incorrect. The algorithm for extracting aspect terms with a rule pattern that contains a dependency relation pair is similar.
    
    In practice, we also construct a dictionary that includes the frequently used aspect terms in the training set. This dictionary is used to extract aspect terms through direct matching. 

	The opinion term extraction rule mining algorithm is similar. But rule patterns related to an opinion word vocabulary are not generated. When extracting opinion terms based on rules, three types of words are considered as possible opinion terms: adjectives, nouns and verbs.
    
    \paragraph{Time Complexity}	Let $L$ be the maximum number of words in an aspect/opinion term, $M$ be the maximum number of words in a sentence, $N$ be the total number of aspect terms in the training set. Then, the time complexity of the rule candidate generation part is $\mathcal{O}(LNM^2)$. There can be at most $LNM^2/T$ candidate rules, so the time complexity of the rule filtering part of the algorithm is $\mathcal{O}(LNM^4/T)$. In practice, the algorithm is fast since the actual number of rule candidates obtained is much less than $LNM^2/T$. 
	
	\subsection{Neural Model}
	
	After the rules are mined, they are applied to a large set of product reviews $D_a$ to obtain the aspect and opinion terms in each sentence. The results are then transformed into BIO tag sequences in order to be used by a neural model. Since the mined rules are inaccurate, there can be conflicts in the results, i.e., a word may be extracted as both an aspect term and an opinion term. Thus, we need two tag sequences for each sentence in $D_a$ to represent the result, one for the aspect terms and the other for the opinion terms. 
	
\begin{figure}
	\centering
	\begin{subfigure}[b]{0.35\textwidth}
		\includegraphics[width=1\linewidth]{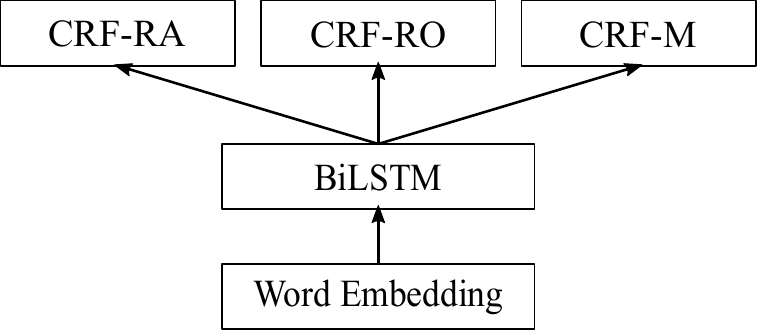}
		\caption{Shared BiLSTM Model.}
		\label{fig:neu1} 
	\end{subfigure}
	
	\begin{subfigure}[b]{0.35\textwidth}
		\includegraphics[width=1\linewidth]{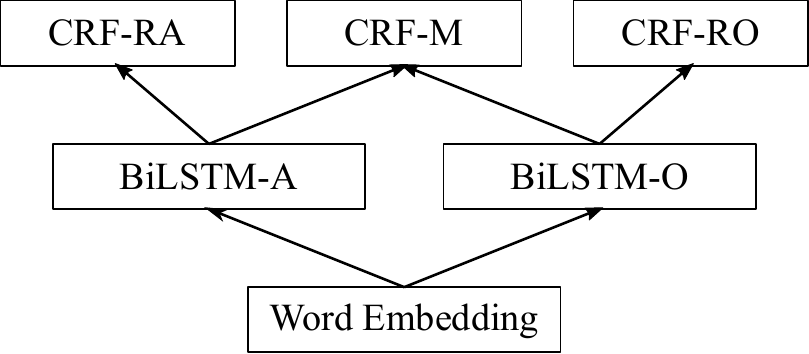}
		\caption{Double BiLSTM Model.}
		\label{fig:neu2} 
	\end{subfigure}
	\caption[]{The structures of two neural aspect and opinion term extraction models. }
	\label{fig:neu}
\end{figure}

Our neural model should be able to learn from the above two tag sequences and a set of manually labeled data. Thus there are three tasks: predicting the terms extracted by the {\it aspect term extraction rules}; predicting the terms extracted by the {\it opinion term extraction rules}; predicting the {\it manual labeling results}. We denote these three tasks as $t_a, t_o$, and $t_m$, respectively. Note that the review sentences in the manually labeled data only need one tag sequence to indicate both aspect terms and opinion terms, since no words in the accurately labeled data can be both an aspect term and an opinion term. 
Then we can train a neural network model with both ground truth supervision and weak supervision. We propose two BiLSTM-CRF \cite{huang2015bidirectional} based models that can be trained based on these three tasks. Their structures are shown in Figure \ref{fig:neu}.
	
We call the model in Figure \ref{fig:neu1} Shared BiLSTM Model and the model in Figure \ref{fig:neu2} Double BiLSTM Model.
Both models use pre-trained embeddings of the words in a sentence as input, then a BiLSTM-CRF structure is used to predict the labels of each word.
They both use three linear-chain CRF layers for the three different prediction tasks: CRF-RA is for task $t_a$; CRF-RO is for task $t_o$; CRF-M is for task $t_m$. 
In Shared BiLSTM Model, the embedding of each word is fed into a BiLSTM layer that is share by the three CRF layers. Double BiLSTM Model has two BiLSTM layers: BiLSTM-A is used for $t_a$ and $t_m$; BiLSTM-O is used for $t_o$ and $t_m$. When they are used for $t_m$, the concatenation of the output vectors of BiLSTM-A and BiLSTM-O for each word in the sequence are used as the input of CRF-M.
	
\paragraph{Training} It is not straightforward how to train these two models. We use two different methods: 1) train on the three tasks $t_a, t_o$ and $t_m$ alternately; 2) pre-train on $t_a$ and $t_o$, then train on $t_m$. In the first method, at each iteration, each of the three tasks is used to update the model parameters for one time. In the second method, the model is first pre-trained with $t_a$ and $t_o$, with these two tasks trained alternately. The resultant model is then trained with $t_m$. We perform early stopping for training. While training with the first method or training on $t_m$ with the second method, early stopping is performed based on the performance (the sum of the $F1$ scores for aspect term extraction and opinion term extraction) of $t_m$ on a validation set. In the pre-training part of the second method, it is based on the sum of the $F1$ scores of $t_a$ and $t_o$. We also add dropout layers \cite{srivastava2014dropout} right after the BiLSTM layers and the word embedding layers.

	\section{Experiments}
	
	This section introduces the main experimental results. We also conducted some experiments related to BERT \cite{devlin2018bert}, which are included in the appendix.
	
	\subsection{Datasets}
	
	We use three datasets to evaluate the effectiveness of our aspect and opinion term extraction approach: SemEval-2014 Restaurants, SemEval-2014 Laptops, and SemEval-2015 Restaurants. They are originally used in the SemEval semantic analysis challenges in 2014 and 2015. Since the original datasets used in SemEval do not have the annotation of the opinion terms in each sentence, we use the opinion term annotations provided by \cite{wang2016recursive} and \cite{wang2017coupled}. Table \ref{tab:dataset-stat} lists the statistics of these datasets, where we use SE14-R, SE14-L, and SE15-R to represent SemEval-2014 Restaurants, SemEval-2014 Laptops, and SemEval-2015 Restaurants, respectively.
	
	\begin{table}
		\begin{center}
			{
				\begin{tabular}{l|c|c|c}
					\hline Dataset & \#Sentences & \#AT & \#OT \\ \hline
					SE14-R (Train) & 3,044 & 3,699 & 3,528 \\
					SE14-R (Test) & 800 & 1,134 & 1,021 \\
					SE14-L (Train) & 3,048 & 2,373 & 2,520 \\
					SE14-L (Test) & 800 & 654 & 678 \\
					SE15-R (Train) & 1,315 & 1,279 & 1,216 \\
					SE15-R (Test) & 685 & 597 & 517 \\
					\hline
				\end{tabular}
			}
		\end{center}
		\caption{\label{tab:dataset-stat} Dataset statistics. AT: aspect terms; OT: opinion terms.}
	\end{table}

Besides the above datasets, we also use a Yelp dataset\footnote{https://www.yelp.com/dataset/challenge} and an Amazon Electronics dataset \cite{he2016ups}\footnote{http://jmcauley.ucsd.edu/data/amazon/} as auxiliary data to be annotated with the mined rules. They are also used to train word embeddings. The Yelp dataset is used for the restaurant datasets SE14-R and SE15-R. It includes 4,153,150 reviews that are for 144,072 different businesses. Most of the businesses are restaurants. The Amazon Electronics dataset is used for the laptop dataset SE14-L. It includes 1,689,188 reviews for 63,001 products such as laptops, TV, cell phones, etc.
	
	\subsection{Experimental Setting}
	For each of the SemEval datasets, we split the training set and use 20\% as a validation set. For SE14-L, we apply the mined rules on all the laptop reviews of the Amazon dataset to obtain the automatically annotated auxiliary data, which includes 156,014 review sentences. For SE14-R and SE15-R, we randomly sample 4\% of the restaurant review sentences from the Yelp dataset to apply the mined rules on, which includes 913,443 sentences. For both automatically annotated datasets, 2,000 review sentences are used to form a validation set, the rest are used to form the training set. They are used while training the neural models of RINANTE. 
	We use Stanford CoreNLP \cite{manning2014stanford} to perform dependency parsing and POS tagging. The frequency threshold integer $T$ in the rule candidate generation part of the rule mining algorithm is set to 10 for all three datasets. The precision threshold $p$ is set to 0.6. We use the same opinion word vocabulary used in \cite{hu2004mining} for aspect term extraction rules. 
	We train two sets of 100 dimension word embeddings with word2vec \cite{mikolov2013distributed} on all the reviews of the Yelp dataset and the Amazon dataset, respectively. The hidden layer sizes of the BiLSTMs are all set to 100. The dropout rate is set to 0.5 for the neural models.
	
	\subsection{Performance Comparison}
	
	To verify the effectiveness of our approach, we compare it with several existing approaches. 
    
	\begin{table*}[t]
		\begin{center}
			{
				\begin{tabular}{l|c|c|c|c|c|c}
                \hline
                & \multicolumn{2}{|c}{SE14-R} & \multicolumn{2}{|c}{SE14-L} & \multicolumn{2}{|c}{SE15-R} \\
					\hline
                    Approach & Aspect & Opinion & Aspect & Opinion & Aspect & Opinion \\ \hline
                    DP \cite{qiu2011opinion} & 38.72 & 65.94 & 19.19 & 55.29 & 27.32 & 46.31 \\
					IHS\_RD \cite{chernyshevich2014ihs} & 79.62 & - & 74.55 & - & - & -  \\
					DLIREC \cite{toh2014dlirec} & 84.01 & - & 73.78 & - & - & -  \\
					Elixa \cite{vicente2017elixa} & - & - & - & - & \textbf{70.04} & -  \\
					WDEmb \cite{yin2016unsupervised} & 84.31 & - & 74.68 & - & 69.12 & - \\
					WDEmb* \cite{yin2016unsupervised} & 84.97 & - & 75.16 & - & 69.73 & - \\
					RNCRF \cite{wang2016recursive} & 82.23 & 83.93 & 75.28 & 77.03 & 65.39 & 63.75 \\
					CMLA \cite{wang2017coupled} & 82.46 & 84.67 & 73.63 & 79.16 & 68.22 & 70.50 \\
					NCRF-AE \cite{zhang2017semi} & 83.28 & 85.23 & 74.32 & 75.44 & 65.33 & 70.16 \\
					HAST \cite{li2018aspect} & 85.61 & - & 79.52 & - & 69.77 & - \\
					DE-CNN \cite{xu2018double} & 85.20 & - & \textbf{81.59} & - & 68.28 & - \\
					\hhline{=|=|=|=|=|=|=}
					Mined Rules & 70.82 & 79.60 & 67.67 & 76.10 & 57.67 & 64.29 \\
					RINANTE (No Rule) & 84.06 & 84.59 & 73.47 & 75.41 & 66.17 & 68.16  \\
					RINANTE-Shared-Alt & \textbf{86.76} & 86.05 & 77.92 & 79.20 & 67.47 & 71.41 \\
					RINANTE-Shared-Pre & 85.09 & 85.63 & 79.16 & 79.03 & 68.15 & 70.44 \\
					RINANTE-Double-Alt & 85.80 & \textbf{86.34} & 78.59 & 78.94 & 67.42 & 70.53 \\
					RINANTE-Double-Pre & 86.45 & 85.67 & 80.16 & \textbf{81.96} & 69.90 & \textbf{72.09} \\
					\hhline{=|=|=|=|=|=|=}
					RINANTE-Double-Pre$^\dagger$ & 86.20 & - & 81.37 & - & 71.89 & - \\
				\hline
				\end{tabular}
			}
		\end{center}
		\caption{\label{tab:perf-all} Aspect and opinion term extraction performance of different approaches. $F1$ score is reported. IHS\_RD, DLIREC, Elixa and WDEmb* use manually designed features. For different versions of RINANTE, ``Shared'' and ``Double'' means shared BiLSTM model and double BiLSTM model, respectively; ``Alt'' and ``Pre'' means the first and the second training method, respectively. RINANTE-Double-Pre$^\dagger$: fine-tune the pre-trained model for only extracting aspect terms, and use the same number of validation samples as DE-CNN \cite{xu2018double}. The results of RINANTE-Double-Pre$^\dagger$ are obtained after this paper gets accepted and are not included in the ACL version. RINANTE-Double-Pre$^\dagger$ achieves the best performance on SE15-R.
		}
	\end{table*}
	
	\begin{itemize}
		\item DP (Double Propagation) \cite{qiu2011opinion}: A rule based approach that uses eight manually designed rules to extract aspect and opinion terms. It only considers noun aspect terms and adjective opinion terms.
		\item IHS\_RD, DLIREC, and Elixa: IHS\_RD \cite{chernyshevich2014ihs} and DLIREC \cite{toh2014dlirec} are the best performing systems at SemEval 2014 on SE14-L and SE14-R, respectively. Elixa \cite{vicente2017elixa} is the best performing system at SemEval 2015 on SE15-R. All these three systems use rich sets of manually designed features.
		\item WDEmb and WDEmb*: WDEmb \cite{yin2016unsupervised} first learns word and dependency path embeddings without supervision. The learned embeddings are then used as the input features of a CRF model. WDEmb* adds manually designed features to WDEmb.
		\item RNCRF: RNCRF \cite{wang2016recursive} uses a recursive neural network model based the dependency parsing tree of a sentence to obtain the input features for a CRF model.
		\item CMLA: CMLA \cite{wang2017coupled} uses an attention based model to get the features for aspect and opinion term extraction. It intends to capture the direct and indirect dependency relations among aspect and opinion terms through attentions.
		Our experimental setting about word embeddings and the splitting of the training sets mainly follows \cite{yin2016unsupervised}, which is different from the setting used in \cite{wang2016recursive} for RNCRF and \cite{wang2017coupled} for CMLA. For fair comparison, we also run RNCRF and CMLA with the code released by the authors under our setting.
		
		\item NCRF-AE \cite{zhang2017semi}: It is a neural autoencoder model that uses CRF. It is able to perform semi-supervised learning for sequence labeling. The Amazon laptop reviews and the Yelp restaurant reviews are also used as unlabeled data for this approach.
		
		\item HAST \cite{li2018aspect}: It proposes to use Truncated History-Attention and Selective Transformation Network to improve aspect extraction.
		
		\item DE-CNN \cite{xu2018double}: DE-CNN feeds both  general-purpose embeddings and domain-specific embeddings to a Convolutional Neural Network model.
	\end{itemize}
	
	We also compare with two simplified versions of RINANTE: directly using the mined rules to extract terms; only using human annotated data to train the corresponding neural model. Specifically, the second simplified version uses a BiLSTM-CRF structured model with the embeddings of each word in a sentence as input. This structure is also studied in \cite{liu2015fine}. We name this approach RINANTE (no rule).
	
	The experimental results are shown in Table \ref{tab:perf-all}.
    From the results, we can see that the mined rules alone do not perform well. However, by learning from the data automatically labeled by these rules, all four versions of RINANTE achieves better performances than RINANTE (no rule). This verifies that we can indeed use the results of the mined rules to improve the performance of neural models. Moreover, the improvement over RINANTE (no rule) can be especially significant on SE14-L and SE15-R. We think this is because SE14-L is relatively more difficult and SE15-R has much less manually labeled training data.
    
    Among the four versions of RINANTE, RINANTE-Double-Pre yields the best performance on SE14-L and SE15-R, while RINANTE-Shared-Alt is slightly better on SE14-R. Thus we think that for exploiting the results of the mined rules, using two separated BiLSTM layers for aspect terms and opinion terms works more stably than using a shared BiLSTM layer. Also, for both models, it is possible to get good performance with both of the training methods we introduce. In general, RINANTE-Double-Pre performs more stable than the other three versions, and thus is suggested to be used in practice. 
    
    We can also see from Table \ref{tab:perf-all} that the rules mined with our rule mining algorithm performs much better than Double Propagation. This is because our algorithm is able to mine hundreds of effective rules, while Double Propagation only has eight manually designed rules. 
    
    Compared with the other approaches, RINANTE (not including RINANTE-Double-Pre$^\dagger$) only fails to deliver the best performance on the aspect term extraction part of SE14-L and SE15-R. On SE14-L, DE-CNN performs better. However, our approach extracts both aspect terms and opinion terms, while DE-CNN and HAST only focus on aspect terms. On SE15-R, the best performing system for aspect term extraction is Elixa, which relies on handcrafted features
    
	\begin{table}
		\begin{center}
			{
				\begin{tabular}{l|c|c|c|c}
					\hline Dataset & \#ATER & \#OTER & \#EAT & \#EOT \\ \hline
					SE14-R & 431 & 618 & 1,453 & 1,205 \\
					SE14-L & 157 & 264 & 670 & 665 \\
					SE15-R  & 133 & 193 & 818 & 578 \\
					\hline
				\end{tabular}
			}
		\end{center}
		\caption{\label{tab:rule-stat} Number of mined rules on each dataset. ATER means aspect term extraction rules; OTER means opinion term extraction rules; EAT and EOT mean the extracted aspect terms and the extracted opinion terms on the corresponding test set, respectively.}
	\end{table}
    
	\begin{table}
		\begin{center}
			{\small
				\begin{tabular}{p{3.4cm}|p{3.2cm}}
					\hline Rule Pattern & Matched Example \\ \hline
					$(nsubj, O, noun*)$ & The \textbf{OS} is great. \\ \hline
					$(amod, noun*, O)$ & Long \textbf{battery life}. \\ \hline
					$(dobj, \text{``has''}, noun*)$ & It has enough \textbf{memory} to run my business. \\ \hline
				$\{(nsubj, VBN, \underline{noun*}),$ $(case, \underline{noun*}, \text{with})\}$ & I am fully satisfied with the \textbf{performance}. \\ \hline
				\end{tabular}
			}
		\end{center}
		\caption{\label{tab:rule-example} Mined aspect extraction rule examples. Shared words in dependency relation pairs are underlined. Aspect terms are in boldface. $O$ matches predefined opinion words; $VBN$ is a POS tag. $noun*$ means the corresponding noun phrase that includes this word should be extracted.}
	\end{table}
    
\subsection{Mined Rule Results}
The numbers of rules extracted by our rule mining algorithm and the number of aspect and opinion terms extracted by them on the test sets are listed in Table \ref{tab:rule-stat}. It takes less than 10 seconds to mine these rules on each dataset on a computer with Intel i7-7700HQ 2.8GHz CPU. The least amount of rules are mined on SE15-R, since this dataset contains the least amount of training samples. This also causes the mined rules to have inferior performance on this dataset. We also show some example aspect extraction rules mined from SE14-L in Table \ref{tab:rule-example}, along with the example sentences they can match and extract terms from. The ``intentions'' of the first, second, and third rules are easy to guess by simply looking at the patterns. As a matter of fact, the first rule and the second rule are commonly used in rule based aspect term extraction approaches \cite{zhuang2006movie,qiu2011opinion}. However, we looked through all the mined rules and find that actually most of them are like the fourth rule in Table \ref{tab:rule-example}, which is hard to design manually through inspecting the data. This also shows the limitation of designing such rules by human beings.
    
    
	\begin{table*}
		\begin{center}
			{\small
				\begin{tabular}{p{4cm}|p{2.7cm}|p{2.4cm}|p{2cm}|p{2cm}}
					\hline Sentence & RINANTE (no rule) & Mined Rules & RINANTE & DE-CNN \\ \hline
					The \textbf{SuperDrive} is quiet. & - & SuperDrive & SuperDrive & SuperDrive \\ \hline
					My life has been enriched since I have been using Apple products.  & life & - & - & - \\ \hline
					It would seem that its \textbf{Mac OS 10.9} does not handle \textbf{external microphones} properly. & Mac OS 10.9 & Mac OS 10.9; microphones & Mac OS 10.9; external microphones & Mac OS 10.9; external microphones \\ \hline
					I love the \textbf{form factor}. & - & form factor & form factor & - \\ \hline
				\end{tabular}
			}
		\end{center}
		\caption{\label{tab:cases} Example sentences and the aspect terms extracted by different approaches. The correct aspect terms are in boldface in the sentences. ``-'' means no aspect terms are extracted.}
	\end{table*}
    
    
\subsection{Case Study}
\label{sec:casestudy}

To help understand how our approach works and gain some insights about how we can further improve it, we show in Table \ref{tab:cases} some example sentences from SE14-L, alone with the aspect terms extracted by RINANTE (no rule), the mined rules, RINANTE (RINANTE-Double-Pre), and DE-CNN. In the first row, the aspect term ``SuperDrive'' can be easily extracted by a rule based approach. However, without enough training data, RINANTE (no rule) still fails to recognize it. In the second row, we see that the mined rules can also help to avoid extracting incorrect terms. The third row is also interesting: while the mined rules only extract ``microphones'', RINANTE is still able to obtain the correct phrase ``external microphones'' instead of blindly following the mined rules. The sentence in the last row also has an aspect term that can be easily extracted with a rule. The result of RINANTE is also correct. But both RINANTE (no rule) and DE-CNN fails to extract it.
	
\section{Conclusion and Future Work}
    In this paper, we present an approach to improve the performance of neural aspect and opinion term extraction models with automatically mined rules. We propose an algorithm to mine aspect and opinion term extraction rules that are based on the dependency relations of words in a sentence. The mined rules are used to annotate a large unlabeled dataset, which is then used together with a small set of human annotated data to train better neural models. The effectiveness of this approach is verified through our experiments. For future work, we plan to apply the main idea of our approach to other tasks.

\section*{Acknowledgments}
This paper was supported by WeChat-HKUST WHAT Lab  and the Early Career Scheme (ECS, No. 26206717) 
from Research Grants Council in Hong Kong. We also thank Intel Corporation for supporting our deep learning related research.

\bibliography{rinante-acl2019}
\bibliographystyle{acl_natbib}

\appendix

\section{Experimental Results with BERT}
\label{sec:bertexp}

We also conduct experiments with the language representation model BERT. The original paper suggests two approaches to apply this model to a sequence labeling problem. The first approach is to fine tune BERT by feeding the final hidden representation for each token into a classification layer; the second approach is to use the hidden layers of the pretrained Transformer as fixed features for a sequence labeling model. They are called the fine-tuning approach and the feature-based approach, respectively. Obviously, the feature-based approach can also be used by RINANTE.

Here, we compare four different appraoches: BiLSTM-CRF + word2vec, BERT fine-tuning, BERT feature-based and RINANTE+BERT. BiLSTM-CRF + word2vec simply uses word2vec embeddins as the input of a BiLSTM-CRF. BERT fine-tuning is the same fine-tuning approach for name entity recognition used in the paper that proposes BERT. BERT feature-based uses the extracted features as the input of a BiLSTM-CRF model. RINANTE+BERT uses our approach RINANTE-DOUBLE-Pre, but with the features extracted by BERT as word embeddings. Both BERT feature-based and RINANTE+BERT use the top four hidden layers as features.

We use the pretrained \textit{BERT-Base, Uncased} model and further pretrain it with the Yelp reviews and Amazon reviews for our restaurant datasets and laptop dataset, respectively. 200-dimensional BiLSTMs are used for both BERT feature-based and RINANTE+BERT.

The experimental results are listed in Table \ref{tab:perf-bert}. We can see that using BERT yields better performance than using word2vec. RINANTE is still able to further improve the performance when contextual embeddings obtained with BERT are used.

\begin{table*}
	\begin{center}
		{
			\begin{tabular}{l|c|c|c|c|c|c}
            \hline
            & \multicolumn{2}{|c}{SE14-R} & \multicolumn{2}{|c}{SE14-L} & \multicolumn{2}{|c}{SE15-R} \\
				\hline
                Approach & Aspect & Opinion & Aspect & Opinion & Aspect & Opinion \\ \hline
				BiLSTM-CRF + word2vec & 84.06 & 84.59 & 73.47 & 75.41 & 66.17 & 68.16  \\
				BERT fine-tuning & 84.36 & 85.50 & 75.67 & 79.75 & 65.84 & 74.21  \\
                BERT feature-based & 85.14 & 85.74 & 76.81 & 81.41 & 66.84 & 73.92 \\
				RINANTE+BERT & \textbf{85.51} & \textbf{86.82} & \textbf{79.93} & \textbf{82.09} & \textbf{68.50} & \textbf{74.54}  \\
			\hline
			\end{tabular}
		}
	\end{center}
	\caption{\label{tab:perf-bert} Aspect and opinion term extraction performance.
	}
\end{table*}

\end{document}